\UseRawInputEncoding 
\documentclass{article}
\usepackage{spconf,amsmath,graphicx}
\usepackage[skip=2pt]{caption}
\usepackage{multirow}
\usepackage[table,xcdraw]{xcolor}
\usepackage{array}
\usepackage{flushend}

\title{Investigation of Training Label Error Impact on RNN-T}
\name{I-Fan Chen, Brian King, Jasha Droppo}
\address{
  Amazon \\
  United States
}
\begin{document}
%
\maketitle
\begin{abstract}
In this paper, we propose an approach to quantitatively analyze impacts of different training label errors to RNN-T based ASR models. The result shows deletion errors are more harmful than substitution and insertion label errors in RNN-T training data. We also examined label error impact mitigation approaches on RNN-T and found that, though all the methods mitigate the label-error-caused degradation to some extent, they could not remove the performance gap between the models trained with and without the presence of label errors. Based on the analysis results, we suggest to design data pipelines for RNN-T with higher priority on reducing deletion label errors. We also find that ensuring high-quality training labels remains important, despite of the existence of the label error mitigation approaches.
\end{abstract}
\section{Introduction}
Automatic speech recognition (ASR) had a significant performance breakthrough \cite{Xiong2017} in the past decade thanks to the deep learning technology and big data \cite{Xiong2017}. The recent trend of all neural based end-to-end ASR such as RNN-T \cite{Graves2012,He} and LAS \cite{Chan}, in the ASR community further leverage the modeling power of neural networks by jointly optimizing ASR components. Such end-to-end approach implies higher demands to training data in the modern ASR systems \cite{Kaplan2020,Henighan2020}. However, human-labeled data process are expensive, and the labeling process is time-consuming. To prepare sufficient training data for data-hungry ASR models today under limited budgets, researchers have explored ways labeling big data, e.g., smartly selecting data for transcription using active learning \cite{Riccardi2005,Drugman}, semi-supervised learning \cite{Drugman,Vesely2013}, and weakly supervised learning \cite{Zhou2018a}. Despite that the approaches alleviate the data amount requirement issue, they may introduce additional label errors to the training data. And models trained with such data may be harmed by those label errors.

The impact of training label errors to neural networks has been well studied in the image research community. Analysis reports \cite{Rolnick2017,Arplt2017} and label error robust training approaches are proposed in the past decade. As the biggest concern for training label errors (or noise, which is an interchangeable term used in this paper) is that neural networks might memorize those errors in the training process. In \cite{Rolnick2017} and \cite{Arplt2017}, both authors found that though neural networks are capable of memorizing error data, the models tend to learn simple patterns shared by multiple training examples first. Based on this finding, \cite{Rolnick2017} proposed that a sufficiently large training set can accommodate a wide range of noise levels in training data. \cite{Arplt2017} further suggests that regularization can differentially hinder memorization to noises in deep neural networks while preserving their ability to learn about real data; it also recommends that higher model capacity is required for models to learn patterns from noisy data. In addition to the suggestions based on the insight of memoization mechanism of neural networks, a number of label error robust training approaches for neural networks were also proposed in  the community. In \cite{Reed2015}, the impact of label errors is mitigated by combining model predicted labels and training labels to form the training target for back propagation. \cite{Goldberger2017,Patrini2017} maintain a noise transition matrix to either adapt the model output or ``correct" the loss value directly. In \cite{Han2018,Yu2019,Jiang2018,Malach2017}, the authors proposed different ways to dynamically select more reliable samples for model training and mitigate the impact of training label errors.


Fewer analyses on training label error impact to ASR models can be found in the ASR community. In \cite{Sundaram}, the authors analyzed impacts of randomly simulated training label error to the conventional hidden Markov model (HMM)-Gaussian Mixture Model (GMM) based acoustic models (AM) and found that insertion errors are more harmful to HMM-GMM based AMs. For label error impact mitigation, most of the proposed methods are based on error transcription filtering \cite{Kurata2011,Wang}. However, in \cite{Dufraux2019}, the authors used a noise transition matrix with beam search to find better alternative transcription to the original error prone version for model training. Note that the above mentioned researches were conducted with $\textless$ 1k hour training data and for models other than the end-to-end RNN-T models.


To understand the impact of label error to RNN-T models, in this paper, we conducted experiments simulating different types of label errors at different scales. Our contributions are the following:
\begin{enumerate}
\item Propose a methodology quantitively analyzing impacts of different types of training label errors on RNN-T ASR systems,
\item Evaluate the effectiveness of existing approaches mitigating training label error impact to RNN-T ASR systems, and
\item Provide insights for designing a training data pipeline handling label error issues that optimizing for RNN-T ASR systems.
\end{enumerate}

\section{Label Error Simulation} \label{ler_sim}
We conducted our analysis on Alexa ASR tasks. To quantitively analyze the impact of label errors to RNN-T models, we simulate training datasets with three types of errors, i.e, deletion, insertion, and substitution, that human transcribers may make. Given most of the Alexa data utterances are short utterances, we restrict at most one label error in a utterance under the scenario where the errors are made un-intentionally by the transcribers. Our label error simulation algorithm is summarized as follows:


Given a target label error rate (LER), $E^{target}$, and a training dataset with $M$ utterances (containing $N$ words), we simulated the label error by
\begin{enumerate}
\item Randomly shuffle the training utterances using a fixed seed to create error injection sequence for the utterances in the original training data. The fixed seed is to ensure the utterances selected for error injection in the later steps are the same for deletion / insertion / substitution datasets generation process, so our experimental results are comparable among different error types.
\item \label{step2} Start from the beginning of the utterance list, inject a simulated label error of the target type (i.e., substitution, insertion, or deletion) to the current utterance. Single-word utterances are skipped to avoid empty transcription after deletion error injection. 
\item \label{step3} Compute the label error rate, $E^{cur}=\frac{C}{N}$, where $C$ is the number of errors injected so far, after the injection.
\item \label{step4} If $E^{cur} \le E^{target}$, go back to step \ref{step2} for next utterance error injection. Otherwise terminate the injection process and use the dataset as the training data with $E^{target}$ LER. 
\end{enumerate}
The subsections below describe the details of how each type of label errors are simulated in this paper.

\begin{table}[h]
	\caption{The label error rates (LER) and their corresponding sentence error rates (SER) used in this paper. The average utterance length for the training dataset is 3.8 words.}
	\label{tab:ler_ser}
  	\centering
	\begin{tabular}{|l|l|l|l|}
	\hline
	\rowcolor[HTML]{EFEFEF} 
	\textbf{LER} & \textbf{1.0 \%} & \textbf{2.0 \%} & \textbf{6.0 \%} \\ \hline
	SER          & 3.8 \%          & 7.6 \%          & 22.8 \%         \\ \hline
	\end{tabular}
\end{table}
\subsection{Deletion Error Simulation}\label{del_sim}
Given an utterance, the deletion error is simulated by randomly removing one non-preserved word from the transcription. We preserve the top 10 most frequent words in the training dataset from being removed as those words are less likely to be missed by transcribers due to their popularity.
\subsection{Insertion Error Simulation}\label{ins_sim}
We simulate the insertion error using a bigram language model (LM) estimated from a matched task dataset.
Given an utterance, we first uniformly sample an insertion position in the utterance. A word sampled from the bigram LM given the context word at the position is then inserted to the position.
\subsection{Substitution Error Simulation}\label{sub_sim}
The substitution errors are simulated based on word pronunciation similarity and language model probability. A soundex \cite{soundex} indexed word lookup table is built from the training vocabulary. The same bi-gram LM in section \ref{ins_sim} is used.

Given an utterance, a substitution error is injected by:

\begin{enumerate}
	\item \label{sub:select_word} Randomly sample a word in the utterance for substitution. To avoid having the substitution errors dominated by wakewords such as `alexa', `echo', and `amazon', where human transcribers unlikely to make mistake with, we consider those words as preserved words in the substitution process and try to avoid selecting them by resampling up to three times if they are selected.
	\item \label{sub:soundex} Compute the soundex of the selected word and use it to look up words with the same soundex in the table. If no other words match the soundex of the selected word, go back to step \ref{sub:select_word} and pick another word for substitution.
	\item \label{sub:substitue} Sample a word from the soundex-matched words based on their n-gram probability at the given utterance position for substitution.
\end{enumerate}

\section{Label Error Impact Mitigation}
Label errors impact and its mitigation is one of a popular topic studied in the image community in the recent decade \cite{Rolnick2017,Arplt2017,Reed2015,Goldberger2017,Patrini2017,Han2018,Yu2019,Jiang2018,Malach2017}. The mitigation approaches usually fall in three categories: i) data based, e.g., data filtering or selection, ii) model capacity based, e.g., increase model or data size, iii) optimization process based: e.g., regularization. In this paper, we evaluate the effectiveness of some of the common approaches on mitigating label error impact to neural networks in these three categories:

\begin{itemize}
	\item \textbf{Dropout}\cite{dropout12}: Dropout is a regularization approach for neural networks by randomly dropping a feature input of a neural network layer with a given probability. It avoids overfitting by forcing the network to not rely on specific input features for decision making. \cite{Arplt2017} has shown that dropout is effective on suppressing neural networks' memorization on label errors without sacrificing the performance on real data.
	\item \textbf{SpecAugment}\cite{spec_aug2019}: SpecAugment is a regularization found useful for all neural end-to-end ASR model training. By partially masking the  spectrogram of the input training utterances on time or frequency axises, the approach forced the neural network to focus on learning global patterns rather than memorizing local patterns by increasing the training data amount.
	\item \textbf{Data Size Increase}: Increase training data size usually lead to better model performance \cite{Kaplan2020}. It may be used to compensate the degradation caused by label errors \cite{Rolnick2017}.
	\item \textbf{Model Size Increase}: Bigger models are more efficient learner for general pattern in the data \cite{Kaplan2020, Arplt2017}. Model size increase thus can be used to mitigate the degradation caused by label errors.
	\item \textbf{Error Utterance Removal}: Removing data samples with errors from training is a straightforward way to reduce label error impact. For ASR, this can be done by removing the utterances with incorrect transcriptions.
\end{itemize}

We analyze the effectiveness of these approaches on mitigating impacts from the three types of label errors to RNN-T models in this paper.

\section{Experiments}
\subsection{Experimental Setup}

All experiments were conducted on de-identified Amazon in-house far-field data. The clean training dataset consists of 5k hour human transcribed data. We used the original clean dataset to create 9 noisy-label datasets at 1\%, 2\%, and 6\% LERs following the procedure explained in section \ref{ler_sim}. All the data are de-identified.

Unless explicitly specified, the encoder of the RNN-T models used in this paper have two LSTM layers with hidden layer dimension of 512, and the decoder has 1 LSTM layer with the same hidden layer dimension. No joint network is used. We use 4k word pieces extracted from the training data with byte-pair-encoding (BPE) criterion as the subword unit to the RNN-T model. 
To train the RNN-T models, we use Adam optimizer with a warm hold decay learning rate scheduler. The learning rate was ramped up from  1e-7 to 5e-4 over 3,000 steps, held there until 100,000 steps, and then exponentially decayed to 1e-5 at 200,000 steps. Each step consisted of a 64 example batch size on a single GPU, while 24 GPUs are used for model training. All the RNN-T models in this paper are trained until convergence. Table \ref{tab:exp_summary} summarizes all the baseline and experimental systems used in this paper.

The evaluation data consists of 30 hours of the de-identified Amazon in-house far-field data. We used beam search with beam size 16 for decoding. Length normalization for search cost is applied to encourage RNN-T model longer output sequences. The maximal number for non-blank expansion was set to 10. Due to Amazon data policy, we could not disclose the absolute word error rate (WER) values; instead, we use relative word error rate (R\_WER), relative substitution error rate (R\_Sub), relative insertion error rate (R\_Ins), and relative deletion error rate (R\_Del) that are relative to the absolute WER value of b0 baseline system for showing the results. The R\_WER for the b0 system is therefore 1.0. And the relation $WER=Sub+Ins+Del$ remains for the relative metrics, i.e., $R\_WER=R\_Sub+R\_Ins+R\_Del$, for each system.

\begin{table*}[t]
\caption{The baseline(b) and experimental(e) systems used in this paper. The table shows training regularization (\textbf{Reg}), label error mitigation approach (\textbf{Miti-Appr}), and the RNN-T \textbf{model size} used for the systems. For all the experimental systems, we use the suffix "\textit{.$\textless$error\_type$\textgreater$$\textless$error\_scale$\textgreater$}" to mark their label error training dataset. In this paper, $\textless$error\_type$\textgreater$ and $\textless$error\_scale$\textgreater$ are sub(substitution)/ins(insertion)/del(deletion) and 1/2/6 \%, respectively. For example, e0.del6 means the experimental RNN-T model trained with the 6\% deletion label error injected training dataset.}
\label{tab:exp_summary}
\begin{tabular}{|c|c|c|c|p{10cm}|}
\hline
\rowcolor[HTML]{EFEFEF} 
{\color[HTML]{333333} 
\textbf{System ID}}      & \textbf{Reg} & \textbf{Miti-Appr}       & \textbf{Model Size} & \textbf{Description}    \\ \hline
  b0                     & N            & \multicolumn{1}{c|}{-}   & x1                  & The original 5k hour data trained RNN-T without any regularization nor label error mitigation approaches. All the R\_\{WER/Sub/Ins/Del\} numbers reported in this paper are relative to the absolute WER value of this RNN-T model. \\ \hline
  e0                     & N            & \multicolumn{1}{c|}{-}   & x1                  & RNN-T models trained with label error injected 5k hour training data. No regularization nor label error mitigation approaches applied. \\ \hline
  b1                     & Y            & \multicolumn{1}{c|}{-}   & x1                  & The original 5k hour data trained RNN-T with regularization.   \\ \hline
  e1                     & Y            & \multicolumn{1}{c|}{-}   & x1                  & RNN-T models trained with label error injected 5k hour training data and regularization.                                                          \\ \hline
  b2                     & Y            & Larger Data              & x1                  & The original 5k, 10k, 20k, 40k, and 80k hour data trained RNN-T models with SpecAugment regularization. \\ \hline
  e2                     & Y            & Larger Data              & x1                  & RNN-T models trained with label error injected 5k, 10k, 20k, 40k, and 80k hour training data and SpecAugment regularization.  \\ \hline
  b3                     & Y            & Larger Model             & x1, x2, x6          & The original 5k hour data trained RNN-T (w/ SpecAugment regularization) at x1, x2, and x6 model size (comparing to b0 model). \\ \hline
  e3                     & Y            & Larger Model             & x1, x2, x6          & RNN-T models trained with label error injected 5k hour training data and SpecAugment regularization at x1, x2, and x6 model size relative to b0 model. \\ \hline
  b4                     & Y            & \multicolumn{1}{c|}{-}   & x2                  & The original 5k hour data trained x2 size RNN-T \\ \hline
  e4                     & Y            & Error utt removal        & x2                  & The x2 size RNN-T models trained by removing the error utterances from the simulated label error injected 5k hour training data. \\ \hline
\end{tabular}
\end{table*}

\subsection{Label Error Impact Analysis} \label{label_error_analysis}
We first analysis the impact of label error types at different scales. Table \ref{tab:exp0} summarizes the experimental results. Both b0 and e0 systems are trained without regularization except early stopping.

We observed that different types of label errors have different impact to RNN-T models. When comparing RNN-T trained with substitution, insertion, and deletion label errors at the same rate, the impact of deletion label error is significantly higher than the other two types of error. For LER=6\%, it is observed deletion label errors cause 9.2\% relative WER degradation (R\_WER from 1.0 to 1.092), while only 3.1\% and 0.4\% relative WER degradations were observed for substitution and insertion label errors respectively. The same trend is observed for smaller LERs. It is also interesting to see that the substitution and insertion label error has little impact for RNN-T performance when LER=1.0\% or 2.0\%. We hypothesize the RNN-T is more sensitive to deletion label error because of the blank output used in RNN-T. As blank can be considered as the probability for a RNN-T model to delete, the deletion label errors directly contribute to the training for the blank output; while for insertion and substitution label errors, their impact is distributed across all non-blank outputs and thus have much less impact than deletion label error.
 
It is also interesting to see with 1.0\% LER for substitution and insertion errors, WER improvements were observed for the RNN-T systems. Possibly because the small amount of injected label errors are playing regularization roles to the RNN-T model, which is similar to label smoothing and penalize the over-confident outputs. We do not see this regularization effect on deletion because, on the opposite, it makes the RNN-T overconfident on emitting blanks.

\begin{table*}[h]
\caption{Impact of substitution, insertion, and deletion training label errors at scale 1\%, 2\%, and 6\% to the RNN-T ASR model performances. R\_WER, R\_Sub, R\_Ins, and R\_Del are the normalized (by the b0 WER) WER, Sub, Ins, and Del error values. We also show relative changes in each ASR error type in the adjacent columns. The degradation caused by deletion label errors is about 3 to 20 times larger than the damage caused by substitution and insertion label errors.}
\label{tab:exp0}
\begin{tabular}{|
>{\columncolor[HTML]{EFEFEF}}l |
>{\columncolor[HTML]{EFEFEF}}c |
>{\columncolor[HTML]{EFEFEF}}l |l|r|l|r|l|r|l|r|}
\hline
{\color[HTML]{333333} \textbf{}} & \multicolumn{2}{l|}{\cellcolor[HTML]{EFEFEF}\textbf{Label Error}}              & \multicolumn{2}{c|}{\cellcolor[HTML]{EFEFEF}\textbf{WER}}                                                              & \multicolumn{2}{c|}{\cellcolor[HTML]{EFEFEF}\textbf{Sub}}                                                              & \multicolumn{2}{c|}{\cellcolor[HTML]{EFEFEF}\textbf{Ins}}                                                              & \multicolumn{2}{c|}{\cellcolor[HTML]{EFEFEF}\textbf{Del}}                                                              \\ \hline
\textbf{}                        & \multicolumn{1}{l|}{\cellcolor[HTML]{EFEFEF}\textbf{Type}} & \textbf{Rate(\%)} & \cellcolor[HTML]{EFEFEF}\textbf{R\_WER} & \multicolumn{1}{p{1.5cm}|}{\cellcolor[HTML]{EFEFEF}\textbf{Chg rel to b0 (\%)}} & \cellcolor[HTML]{EFEFEF}\textbf{R\_Sub} & \multicolumn{1}{p{1.5cm}|}{\cellcolor[HTML]{EFEFEF}\textbf{Chg rel to b0 (\%)}} & \cellcolor[HTML]{EFEFEF}\textbf{R\_Ins} & \multicolumn{1}{p{1.5cm}|}{\cellcolor[HTML]{EFEFEF}\textbf{Chg rel to b0 (\%)}} & \cellcolor[HTML]{EFEFEF}\textbf{R\_Del} & \multicolumn{1}{p{1.5cm}|}{\cellcolor[HTML]{EFEFEF}\textbf{Chg rel to b0 (\%)}} \\ \hline
\textbf{b0}                      & \textbf{N/A}                                               & \textbf{0.0}      & 1.000                                   & 0.0\%                                                                        & 0.603                                   & 0.0\%                                                                          & 0.141                                   & 0.0\%                                                                        & 0.256                                   & 0.0\%                                                                        \\ \hline
\textbf{e0.sub1}                 & \cellcolor[HTML]{EFEFEF}                                   & \textbf{1.0}      & 0.992                                   & -0.8\%                                                                       & 0.598                                   & -0.8\%                                                                       & 0.131                                   & -7.1\%                                                                       & 0.263                                   & 2.7\%                                                                        \\ \cline{1-1} \cline{3-11} 
\textbf{e0.sub2}                 & \cellcolor[HTML]{EFEFEF}                                   & \textbf{2.0}      & 1.007                                   & 0.7\%                                                                        & 0.610                                   & 1.2\%                                                                        & 0.143                                   & 1.4\%                                                                        & 0.254                                   & -0.8\%                                                                       \\ \cline{1-1} \cline{3-11} 
\textbf{e0.sub6}                 & \multirow{-3}{*}{\cellcolor[HTML]{EFEFEF}\textbf{Sub}}     & \textbf{6.0}      & 1.031                                   & 3.1\%                                                                        & 0.626                                   & 3.8\%                                                                        & 0.143                                   & 1.4\%                                                                        & 0.262                                   & 2.3\%                                                                        \\ \hline
\textbf{e0.ins1}                 & \cellcolor[HTML]{EFEFEF}                                   & \textbf{1.0}      & 0.997                                   & -0.3\%                                                                       & 0.593                                   & -1.7\%                                                                       & 0.137                                   & -2.8\%                                                                       & 0.267                                   & 4.3\%                                                                        \\ \cline{1-1} \cline{3-11} 
\textbf{e0.ins2}                 & \cellcolor[HTML]{EFEFEF}                                   & \textbf{2.0}      & 1.004                                   & 0.4\%                                                                        & 0.595                                   & -1.3\%                                                                       & 0.129                                   & -8.5\%                                                                       & 0.280                                   & 9.4\%                                                                        \\ \cline{1-1} \cline{3-11} 
\textbf{e0.ins6}                 & \multirow{-3}{*}{\cellcolor[HTML]{EFEFEF}\textbf{Ins}}     & \textbf{6.0}      & 1.004                                   & 0.4\%                                                                        & 0.587                                   & -2.7\%                                                                       & 0.137                                   & -2.8\%                                                                       & 0.280                                   & 9.4\%                                                                        \\ \hline
\textbf{e0.del1}                 & \cellcolor[HTML]{EFEFEF}                                   & \textbf{1.0}      & 1.026                                   & 2.6\%                                                                        & 0.598                                   & -0.8\%                                                                       & 0.124                                   & -12.1\%                                                                      & 0.305                                   & 19.1\%                                                                       \\ \cline{1-1} \cline{3-11} 
\textbf{e0.del2}                 & \cellcolor[HTML]{EFEFEF}                                   & \textbf{2.0}      & 1.036                                   & 3.6\%                                                                        & 0.570                                   & -5.5\%                                                                       & 0.106                                   & -24.8\%                                                                      & 0.360                                   & 40.6\%                                                                       \\ \cline{1-1} \cline{3-11} 
\textbf{e0.del6}                 & \multirow{-3}{*}{\cellcolor[HTML]{EFEFEF}\textbf{Del}}     & \textbf{6.0}      & 1.092                                   & 9.2\%                                                                        & 0.510                                   & -15.4\%                                                                      & 0.070                                   & -50.4\%                                                                      & 0.511                                   & 99.6\%                                                                       \\ \hline
\end{tabular}
\end{table*}

\subsection{Mitigation Analysis: Regularization}
We test label error mitigation approaches on RNN-T systems trained with LER 6\% data. In this experiment, we compare the effectiveness of dropout \cite{dropout12} and SpecAugment \cite{spec_aug2019} on mitigating the training label error impact for RNN-T. An additional set of baselines b1 consists of RNN-T models trained with clean label data and both regularization approaches are also built for comparison. Table \ref{tab:exp1} summarizes the results.

Both dropout and SpecAugment improve the RNN-T models trained with label error. The R\_WER values of e1 systems in Table \ref{tab:exp1} for both regularization approaches are below 1.0, showing all the systems are better than the b0 baseline on WER. However, the remaining gap between b1 and e1 systems indicates the mitigation of label errors impact from regularization is limited for RNN-T. We also observed that SpecAugment provides better R\_WER improvement over dropout; and combining the two does not provide further improvement. Based on this result, for all the following experiments, we applied SpecAugment regularization in system training.

\begin{table}[h]
	\caption{Comparison between dropout and SpecAugment on mitigating label error impact to RNN-T. The values in the No Reg (no regularization) column are from Table \ref{tab:exp0}. Both regularization approaches help, though spec augmentation shows larger improvement. We also noted that combining the two does not provide additional gains.}
	\label{tab:exp1}
\begin{tabular}{|
>{\columncolor[HTML]{EFEFEF}}c |c|c|c|p{1.5cm}|}
\hline
\textbf{R\_WER}  & \cellcolor[HTML]{EFEFEF}\textbf{No Reg} & \cellcolor[HTML]{EFEFEF}\textbf{Dropout} & \cellcolor[HTML]{EFEFEF}\textbf{SpecAug} & \cellcolor[HTML]{EFEFEF}\textbf{Dropout+ SpecAug} \\ \hline
\textbf{b1}      & 1.000                                   & 0.924                                    & 0.777                                    & -                                                  \\ \hline
\textbf{e1.sub6} & 1.031                                   & 0.940                                    & 0.804                                    & 0.972                                              \\ \hline
\textbf{e1.ins6} & 1.004                                   & 0.929                                    & 0.779                                    & 0.965                                              \\ \hline
\textbf{e1.del6} & 1.092                                   & 0.992                                    & 0.862                                    & 1.024                                              \\ \hline
\end{tabular}
\end{table}

\subsection{Mitigation Analysis: Training Data Size Increase}
\begin{figure}[h]
	\centering
	\includegraphics[width=\linewidth]{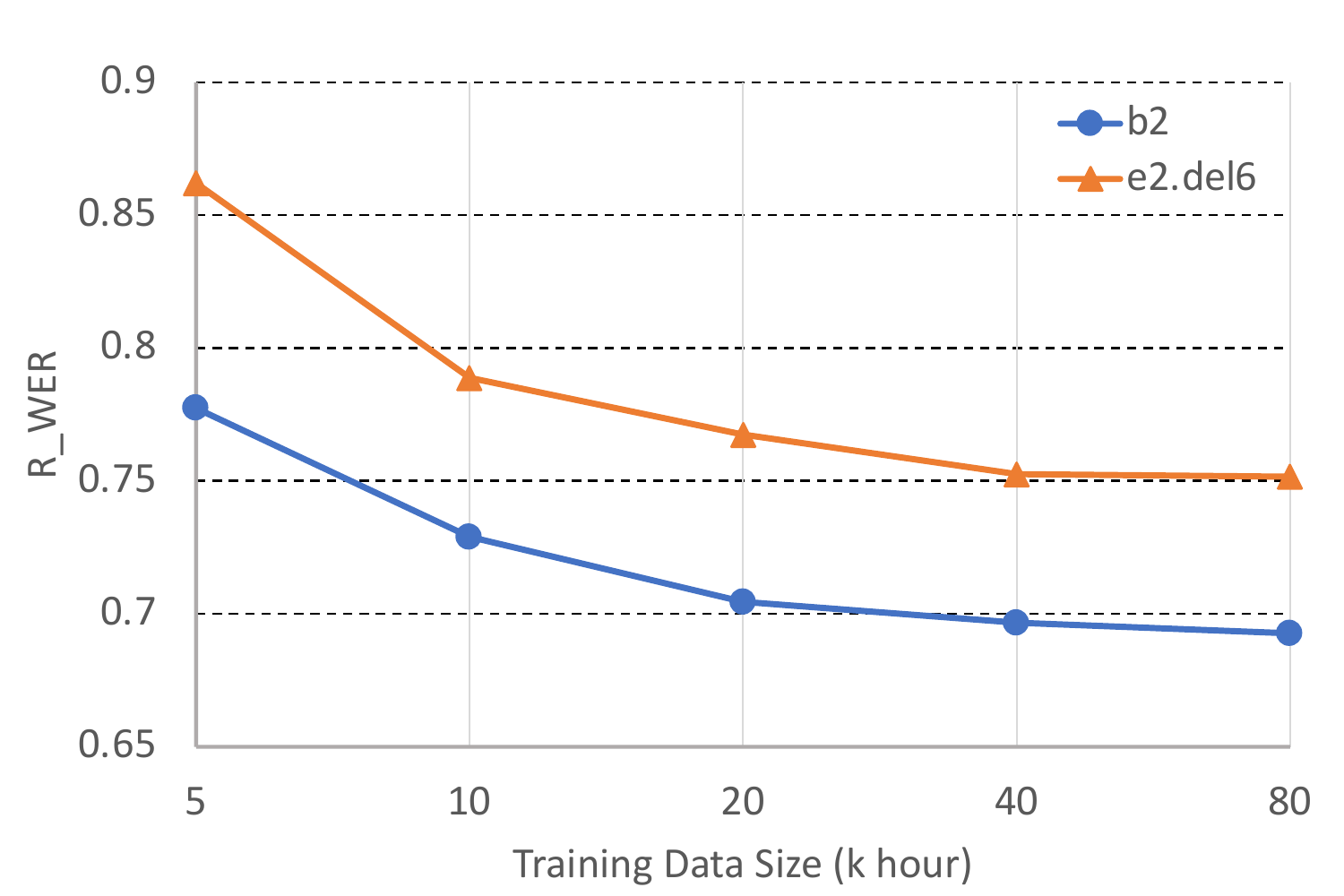} 
	\caption{R\_WER curve for RNN-T models trained with no training label errors vs. 6\% deletion errors using 5, 10, 20, 40, and 80 k hours of training data. The gap  between the asymptotes of the two curves shows training data increase could not mitigate the harm of deletion label errors.}
	\label{fig:exp2}
\end{figure}

In this experiment, we tried to understand whether label error impacts can be mitigated by increasing training data amount. We trained RNN-T models with 10k, 20k, 40k, and 80k hours of Alexa data in addition to the original 5k hour dataset to evaluate the relation between training data size and RNN-T model performance with the presence of label errors. For comparison, we trained two sets of RNN-T models: baseline b2 systems are the RNN-T models trained on clean label data, while e2.del6 systems are the models trained with simulated deletion errors at LER=6\%. Both set of the RNN-T models are trained with SpecAugment regularization. Figure \ref{fig:exp2} summarizes the results.

From Figure \ref{fig:exp2}, we see increasing training data size improves R\_WER for the e2.del6; however similar improvement is also observed for the b2 baseline, and the R\_WER curves of the b2 and e2.del6 converged to different asymptotes with the same performance gap at 5k hour. 

The observation is different from \cite{Rolnick2017}, where the authors found that, given sufficient data, the neural networks reach similar results for data with different error levels (Figure 8 in \cite{Rolnick2017}). We hypothesize, besides the model and task differences\footnote{The observation of \cite{Rolnick2017} is based on ConvNet models on image classification tasks}, such observation difference is due to the asymmetry between label error contribution on the blank and the non-blank RNN-T outputs \footnote{For RNN-T, deletion label errors majorly cause the increase of blank output probability after training; while the impact of insertion and substitution errors distribute across all the non-blank outputs, e.g., the 4k word-piece outputs in this paper.}. Due to asymmetry, the impact of label error does not cancel out even when training data size being significantly increased. The result suggests that, for RNN-T, performance degradation caused by label errors (especially  deletion errors) may not go away by simply increasing training data size.

\subsection{Mitigation Analysis: Model Size Increase}\label{sec:mdl_sz_inc}

The RNN-T models in the experiments so far use the model structure of a 2-layer LSTM encoder and a 1-layer LSTM decoder. We consider the model structure as the base structure in this paper and annotate its size as x1. The total number of parameters for this base structure is about 12 million. To understand the relation between model size and label error impact, we increase the model size by increasing the LSTM layer numbers and dimensions in encoders and decoders. We created two additional model structures with the total parameters 2 and 6 times to the original structure, annotated as x2 and x6. Details are summarized in Table \ref{tab:param_num}. All the models are trained with SpecAugment regularization on the same 5k hour datasets, including the original data with clean labels (b3), and 5k hour data with LER=6\% injected substitution (e3.sub6), insertion (e3.ins6), and deletion (e3.del6).

\begin{figure}[h]
	\centering
	\includegraphics[width=\linewidth]{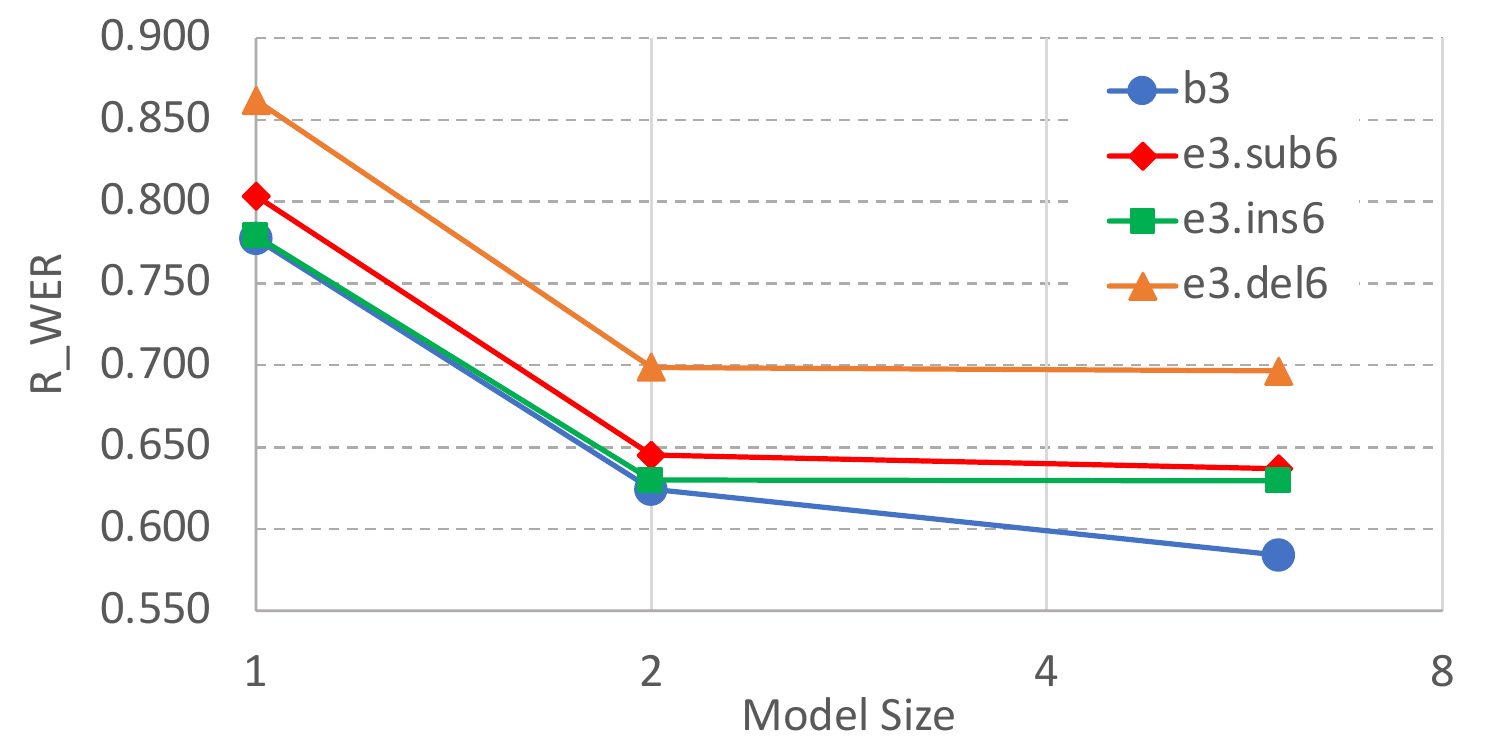} 
	\caption{R\_WER curve for RNN-T models with x1, x2, and x6 model size to the base structure.}
	\label{fig:exp3}
\end{figure}

Figure \ref{fig:exp3} shows the relation between R\_WER and model size. Increasing model size from x1 to x2 provided significant improvement for all the systems. This confirms the observation from \cite{Arplt2017} that larger model capacity is required for achieving same performance when label error present. We also the found the diminishing returns of model size increase occurs earlier when label errors are present. For e3.sub6, e3.ins6, and e3.del6, no further R\_WER improvements\footnote{We do not see degradation thanks to early stop} are observed when increasing size from x2 to x6; while we still see R\_WER improvement for b3 in Figure \ref{fig:exp3}. The result shows that despite larger models performing better in general, that improvement is limited by data quality.

\begin{table}[h]
	\caption{The structures and parameter numbers of the RNN-T models used in the investigation of model size increase impact to label errors.}
	\label{tab:param_num}
	\centering
	\begin{tabular}{|l|l|l|l|}
		\hline
		Model size       & encoder & decoder & \# Param   \\ \hline
		x1       & 2x512   & 1x512   & 11,795,778 \\ \hline
		x2  & 5x512   & 2x512   & 20,192,578 \\ \hline
		x6  & 5x1024  & 2x1024  & 67,640,130 \\ \hline
	\end{tabular}
\end{table}

\subsection{Mitigation Analysis: Oracle Error Utterances Filtering}

Removing training utterances containing label errors is the most straightforward way to ensure training data quality with a trade-off on losing a portion of the training data. We evaluated the effectiveness of such approach under an ideal scenario where it is known which utterances have errors. We focused on the models trained with LER=6\%. From Table \ref{tab:ler_ser}, this means when removing all the utterances with label errors from LER=6\% dataset, we would only have 77.2\% of the training utterances remaining for model training.

We trained e4.sub6, e4.ins6, and e4.del6 using the error-removed simulation datasets with the x2 model structure and compared the performance with the e3 (x2) models in section \ref{sec:mdl_sz_inc}. Table \ref{tab:utt_removal} summarizes the results. For RNN-T, the degradation caused by training size reduction is more severe than the degradation introduced by 6\% LER substitution and insertion errors. The trade-off between training size and label errors is only worthwhile for deletion errors.

\begin{table}[h]
	\caption{R\_WER for RNN-T models trained with LER=6\% dataset (e3) and models trained with the training set removing the error utterances (e4). Both e3 and e4 are with x2 structure from Table \ref{tab:param_num}. The degradation caused by training size reduction is more severe than the degradation introduced by 6\% LER substitution and insertion errors. For RNN-T, the trade-off between training size and label errors is only worthy for deletion label errors.}
	\label{tab:utt_removal}
	\centering
	\begin{tabular}{|l|p{1.2cm}|l|l|l|l|}
		\hline
		             & \textbf{Remove Err Utt} & \textbf{R\_WER} & \textbf{R\_Sub} & \textbf{R\_Ins} & \textbf{R\_Del} \\ \hline
		e3.sub6      & N & 0.645  & 0.397  & 0.091  & 0.157  \\ \hline
		e4.sub6      & Y & 0.683  & 0.412  & 0.093  & 0.179  \\ \hline
		e3.ins6      & N & 0.630  & 0.377  & 0.097  & 0.156  \\ \hline
		e4.ins6      & Y & 0.684  & 0.414  & 0.093  & 0.178  \\ \hline
		e3.del6      & N & 0.699  & 0.336  & 0.055  & 0.308  \\ \hline
		e4.del6      & Y & 0.697  & 0.422  & 0.093  & 0.182  \\ \hline
\end{tabular}
\end{table}

\section{Conclusion}
In this paper, we quantitatively analyzed impacts of different types of label errors to RNN-T models on ASR tasks. The results suggested deletion label errors are much more harmful than substitution and insertion errors for RNN-T. We hypothesize this is because the distribution asymmetry between blank and non-blank RNN-T outputs. The result suggests labeling procedures for RNN-T should treat these three types of label error with different priority and focus on reducing deletion label errors. We examined mitigation approaches for label errors and found that, though all the methods mitigate the impact to some extent, they could not remove the performance gap between clean-trained and label-error data trained models. However, we acknowledge the mitigation approaches we examined in this paper use fixed training data, and for self-learning approaches where training sets are dynamically updated in the training process the observation might be different. We leave this as our future work for investigation.


%

%

\bibliographystyle{IEEEbib}
\bibliography{library}

\end{document}